\documentclass[11pt]{article}
\usepackage{coling2018}
\usepackage{times}
\usepackage{url}
\usepackage{latexsym}
\usepackage{amsmath}
\usepackage{multirow}
\usepackage[T1]{fontenc}
\usepackage{lmodern}

\usepackage{hyperref}

\title{HausaMT v1.0: Towards English--Hausa Neural Machine Translation}

\author{Adewale Akinfaderin \\
 Data Duality\\
  {\tt waleakinfaderin@gmail.com} 
 \\}

\date{}

\begin{document}
\maketitle
\begin{abstract}
Neural Machine Translation (NMT) for low-resource languages suffers from low performance because of the lack of large amounts of parallel data and language diversity. To contribute to ameliorating this problem, we built a baseline model for English--Hausa machine translation, which is considered a task for low--resource language. The Hausa language is the second largest Afro--Asiatic language in the world after Arabic and it is the third largest language for trading across a larger swath of West Africa countries, after English and French. In this paper, we curated different datasets containing Hausa--English parallel corpus for our translation. We trained baseline models and evaluated the performance of our models using the Recurrent and Transformer encoder--decoder architecture with two tokenization approaches: standard word--level tokenization and Byte Pair Encoding (BPE) subword tokenization.
\end{abstract}

\section{Introduction}
Hausa is a language spoken in the western part of Africa. It belongs to the Afro--Asiatic phylum and it is the second most spoken native language on the continent, after Swahili. The language is spoken by more than 40 million people as a first language and about 15 million people use it as a second and third language. Most of the speakers are concentrated in Nigeria, Niger and Chad -- all resulting to both anglophone and francophone influences \cite{Sab:18,Eber:19}. Our work on curating datasets and creating evaluation benchmark for English--Hausa Neural Machine Translation (NMT) is inspired by the socio--linguistic facts of the Hausa language. Hausa has been referred to as the largest internal political unit in Africa. There has been extensive linguistic academic research on Hausa and the language benefits from the existence of trans--border communication in the West African Sahel belt and the availability of international radio stations like the BBC Hausa and Voice of America Hausa \cite{Odo:13}.

The exponential growth of social media platforms have eased communication among users. However, the advances in technological adoption have also informed the need to translate human languages. In low--resource countries, the language inequality can be ameliorated by using machine translation to bridge gaps in technological, political and socio--economic advancements~\cite{Odo:16}. The recent successes in NMT over Phrased--Based Statistical Machine Translation (PBSMT) for high--resource data conditions can be leveraged to explore best practices, data curation and evaluation benchmark for low--resource NMT tasks~\cite{Ben:16,Isa:17}. Using the JW300, Tanzil, Tatoeba and Wikimedia public datasets, we trained and evaluated baseline NMT models for Hausa language.

\blfootnote{4th Widening NLP Workshop, Annual Meeting of the Association for Computational Linguistics, ACL 2020}

\section{Related Works}

\paragraph{\emph{Hausa Words Embedding}:}
Researchers have recently curated datasets and trained word embedding models for the Hausa language. The results from this trained models have been promising, with approximately 300\% increase in prediction accuracy over other baseline models ~\cite{Abd:19}.
 
\paragraph{\emph{Masakhane}:}
Due to the linguistic complexity and morphological properties of languages native to continent of Africa, using abstractions from successful resource--rich cross--lingual machine translation tasks often fail for low--resource NMT task. The Masakhane project was created to bridge this gap by focusing on facilitating open--source NMT research efforts for African languages~\cite{Ori:20}.

\section{Dataset Description}
For the HausaMT task, we used the JW300, Tanzil, Tatoeba and Wikimedia public datasets. The JW300 dataset is a crawl of the parallel data available on Jehovah Witness' website. Most of the data are from the magazines, \emph{Awake!} and \emph{Watchtower}, and they cover a diverse range of societal topics in a religious context~\cite{Agi:19}. The Tatoeba database is a collection of parallel sentences in 330 languages~\cite{Rai:18}. The dataset is crowdsourced and published under a Creative Commons Attribution 2.0 license. The Tanzil dataset is a multilingual text aimed at producing a highly verified multi-text of the Quran text~\cite{Zar:07}. The Wikimedia dataset are parallel sentence pairs extracted and filtered from noisy parallel and comparable wikipedia copora~\cite{Wol:14}. For this work, we trained on two datasets which are: 1) the JW300 as our baseline, and 2) All the datasets combined. The number of tokens, number of sentences and statistical properties of the datasets are in Table \ref{table:results1}.

\begin{table}[!h]
\centering
\begin{tabular}{c|c|c|c|c|c}
\hline
{\textbf{Dataset}} & \multicolumn{2}{c|}{\textbf{Sentence Length (Mean $\pm$ Std)}} & \multicolumn{2}{c|}{\textbf{Tokens}}   & {\textbf{Sentences}} \\ \cline{2-5} & English     & Hausa    & English & Hausa &  \\ \hline
JW300                                & 18.11 $\pm$ 10.53                 & 20.14 $\pm$ 11.57              & 4,051,322                 & 4,506,787               & 223,723                                         \\ \hline
All$^{*}$ & 19.71 $\pm$ 24.31                 & 21.28 $\pm$ 24.60              & 6,919,805                 & 7,471,256             & 351,024                                          \\ \hline
\end{tabular}
\caption{Dataset summary.$^{*}$Combination of JW300, Tanzil, Tatoeba and Wikimedia datasets. }
\label{table:results1}
\end{table}

\section{Experiments and Results}

For our baseline model, we trained a recurrent-based model with the Long Short-Term Memory (LSTM) network as our encoder and decoder type with the Luong attention mechanism~\cite{Luo:15}. To achieve an improved benchmark, we also trained a Transformer encoder--decoder model. The Transformer is based on attention mechanism and the training time is significantly faster than architectures based on convolutional or recurrent networks~\cite{Vas:17}. For the hyperparameters used to train both the recurrent and transformer based architecture, we used an embedding size of 256, hidden units of 256, batch size of 4096 and an encoder and decoder depth of 6 respectively. 

\begin{table}[!h]
\centering
\begin{tabular}{c|c|cc|cc}
\hline
\multirow{2}{*}{\textbf{Dataset}} & \multirow{2}{*}{\textbf{Model}} & \multicolumn{2}{c}{\textbf{BPE}} & \multicolumn{2}{c}{\textbf{Word}} \\ \cline{3-6} 
 &  & \textit{dev} & \textit{test} & \textit{dev} & \textit{test} \\ \hline
\multirow{2}{*}{JW300} & Recurrent & 20.06 & 19.39 & 25.36 & 24.75 \\ 
 & Transformer & \textbf{21.33} & \textbf{20.38} & \textbf{28.71} & \textbf{28.06} \\ \hline
\multirow{2}{*}{All} & Recurrent & 31.89 & 33.48 & 40.78 & 42.29 \\  
 & Transformer & \textbf{31.91} & 32.42 & \textbf{44.42} & \textbf{45.98} \\ \hline
\end{tabular}
\caption{BLEU scores for BPE and word-level tokenization. Best scores of the Transformer model against the Recurrent are highlighted in bold.}
\label{table:results2}
\end{table}

To preprocess the parallel corpus, we used the standard word--level tokenization and Byte Pair Encoding (BPE)~\cite{Gag:94}. The BPE is a subword tokenization which has become a successful choice in translation tasks. The model was trained based on the 4000 BPE tokens used on a recent machine translation study for South African languages~\cite{Mar:19}. To train our model, the Joey NMT minimalist toolkit, which is open source and based on PyTorch was used~\cite{Kre:19}. The models were trained using a Tesla P100 GPU. The model training for the baseline and repeated tasks (datasets \& tokenization type) took between 5-9 hours for each run.

\section{Conclusion and Future Work}
Evaluating the model on the test set, we observed that the word--level tokenization outperform the BPE by a BLEU score factor of \textasciitilde1.27--1.42 times (Table \ref{table:results2}). The qualities of the English to Hausa translations using both word--level and BPE subword tokenizations were rated positively by first language speakers. Table \ref{table:results3} shows some of the translations example.

\begin{table}[!h]
\centering
\begin{tabular}{ll}
\hline
\hline

\textbf{Source:} & This is normal, because they themselves have not been anointed. \\ 
\textbf{Reference:} & Hakan ba abin mamaki ba ne don ba a shafa su da ruhu mai tsarki ba. \\ 
\textbf{Hypothesis:} & Wannan ba daidai ba ne, domin ba a shafe su ba. \\ \hline
\textbf{Source:} & A white - haired man in a frock coat appears on screen. \\ 
\textbf{Reference:} & Wani mutum mai furfura ya bayyana da dogon kwat a majigin. \\ 
\textbf{Hypothesis:} & \begin{tabular}[c]{@{}l@{}}Wani mutum mai suna da wani mutum mai suna da ke cikin mota yana\\  da nisa a cikin kabari\end{tabular} \\ \hline
\textbf{Source:} & Why is that of vital importance? \\ 
\textbf{Reference:} & Me ya sa hakan yake da muhimmanci? \\ 
\textbf{Hypothesis:} & Me ya sa wannan yake da muhimmanci? \\ \hline \hline
\end{tabular}
\caption{Example Translations.}
\label{table:results3}
\end{table}

A significant portion of both the training and test datasets are from the the JW300 parallel data, which are religious texts. We acknowledge that for us to reach a viable state of real-world translation quality, we need to evaluate our model on "general" Hausa data. However, parallel data for other out-of-domain areas does not exist. High-yielding avenue for future work include evaluating on English texts in other domains and crowd-sourcing L1 speakers to manually evaluate the quality of the translations by editing. The post edited translation can then be used as the reference to calculate the evaluation metric. Other future work include carrying out an empirical study to explore the effect of word--level and subword tokenizations. Other methods such as the linguistically motivated vocabulary reduction (LMVR) have shown to perform better for languages in the Afro--Asiatic family~\cite{Ata:18}. The datasets, pre--trained models, and configurations are available on Github.\footnote{https://github.com/WalePhenomenon/Hausa-NMT}

\section*{Acknowledgements}

The author would like to thank Gabriel Idakwo for the qualitative analysis of the translations.


\end{document}